# Analogy-Based and Case-Based Reasoning: Two sides of the same coin

**Michael Gr. Voskoglou** [1] , **Abdel-Badeeh M. Salem** [2]

[1] School of Technological Applications
Graduate Technological Educational Institute of Western Greece
mvosk@hol.gr , http://eclass.teipat.gr/eclass/courses/523102

[2] Faculty of Computer & Information Sciences
Ain Shams University, Cairo, Egypt
absalem@cis.asu.edu.eg , abmsalem@yahoo.com

### Abstract

*Analogy-Based (or Analogical) and Case-Based Reasoning (ABR and CBR) are two similar problem solving processes based on the adaptation of the solution of past problems for use with a new analogous problem. In this paper we review these two processes and we give some real world examples with emphasis to the field of Medicine, where one can find some of the most common and useful CBR applications. We also underline the differences between CBR and the classical rule-induction algorithms, we discuss the criticism for CBR methods and we focus on the future trends of research in the area of CBR.*

## 1. Introduction

As the world economy moved from an industrial to a knowledge economy, it can be argued that the nature of many problems also changed and new problems have arisen which may require a different approach to overcome them. Educational institutions and governments have recognized long ago the importance of Problem Solving (PS) and volumes of research have been written about it (e.g. [28], [60], etc). Universities and other higher learning institutions are entrusted with the task of producing graduates that have better PS skills among other higher order thinking skills ([11], [64], etc).

Mathematics by its nature is a subject whereby PS forms its essence. In earlier papers [114-115] we have examined the role of the problem in learning mathematics and we have attempted a review of the evolution of research on PS in mathematics education from its emergency as a self sufficient science until today. The above research started during the period of 1950's and 1960's and it was based on Polya's ideas on the use of the heuristic strategies in PS [67-71]

One of the most important of these strategies is the strategy of the *analogous problem*: When the solver is not sure of the appropriate procedure to solve a given problem (called the *target problem*), a good hint would be to look for a similar problem solved in the past, and then try to adapt the solution procedure of this problem for use with the target problem. The important benefit of this strategy is that it precludes the necessity of constructing a new solution procedure.

Using the above strategy one has to specify it according to the form of the target problem; e.g. to solve a complex problem with many variables he/she may consider first an analogous problem with fewer variables, to solve a geometric problem in space he/she may consider first the corresponding problem in the plane, etc ; see also [69].

In a more general context (not only for mathematics) *Analogy-Based Reasoning* (ABR) or *Analogical Reasoning* is the process of solving new problems based on the solutions of similar past problems. However this strategy can be difficult to implement in PS, because it requires the solver to attend to information other than the problem to be solved. Thus the solver may come up empty-handed, either because he/she has not solved any similar problems in the past, or because he/she fails to realize the relevance of previous problems. But, even if an analogue is retrieved, the solver must know how to use it to determine the solution procedure for the target problem.

A characteristic example is the experiment of Gick and Holyoak [25-26] on the known as the Dunker's [19] tumor problem: You have a patient with an inoperable stomach tumor. There are some rays that, at sufficient intensity, destroy organic tissue. How can you free the patient of the tumor without destroying the healthy tissue surrounding it?

The desired solution is to use a system of multiple machines to emit low-intensity rays from different directions. These rays will converge on the tumor and their combined effect will destroy it.

In first case only a 10% of the subjects gave the correct solution. Next, before presenting the problem to another group of subjects, it was given to read an analogous story about a general, who wants to capture a fortress, and he is able to do so by sending parts of his army down each of several roads, all of which converge on the fortress. In this case the percentage of the correct solutions was increased to the 30% of the subjects, while a further spectacular increase to 70% happened, when subjects were given the hint to use the story above for the solution of the target problem.

We must finally point out that, the application of this strategy may lead sometimes to false conclusions (see section 2.2: *negative transfer*). This usually happens, when emphasis is given to the surface and not to the structural (solution relevant) characteristics of the target problem (e.g. see [21]). Thus, according to Bazzini [12], analogy is recognizable as a double edged weapon: as means to generate new knowledge and as a potential source of misconceptions.

The importance of ABR in human thinking has been recognized years ago. In fact, there is a considerable number of studies developed and many experiments performed on individuals by mathematicians, psychologists and other scientists about the ABR process (see section 2.2).

However, it is the *Case-Based Reasoning* (CBR) approach to PS and learning (for computers and people) that has got a lot of attention over the last few years, because as an intelligent-systems method enables information managers to increase efficiency and reduce cost by substantially automating processes such as diagnosis, scheduling and design (see section 3.5). The term PS is used in this case in a wide sense, coherent with common practice within the area of knowledge-based systems in general. This means that it is not necessarily the finding of a concrete solution to an application problem, it may be any problem put forth by the user. For example, to justify or criticize an already proposed solution, to interpret a problem situation, to generate a set of possible solutions, or generate explanations in observable data, are also PS situations.

Notice that the term ABR is sometimes used as a synonymous of the typical CBR approach [110]. However is often used also to characterize methods, that solve new problems based on past cases of *different domains* ([27], [39]), while typical CBR methods focus on single-domain cases (a form of intra-domain analogy).

In the present paper we review these two similar PS methods (ABR and CBR) and we present some examples of their applications in practice with emphasis to the field of Medicine, where one can find some of the most common and useful CBR applications. We also focus on the future trends of research in CBR, while in our conclusions' section we underline the differences between CBR and the classical rule-induction algorithms and we discuss the criticism on CBR methods.

## 2. Analogy- Based Reasoning

### 2.1 Transfer of knowledge

Solution of problems by analogy is a special case of the general class of the *transfer of knowledge*, i.e. of the use of already existing knowledge to produce new knowledge. Despite the centrality of transfer to teaching and learning it is only recently that the nature of the transfer process has received detailed analysis.

According to Voss [121] any instance of acquisition of knowledge involves the use of existing knowledge, therefore learning is a specific case of the general class of transfer and so it can be seen as subordinate to transfer. When placed in this relationship with learning, transfer takes a level of complexity considerably greater than that of a simple extension of learning resulting from generalization [22]. This involves efficient execution of awareness, schema induction and automation of problem operators [18]. Salomon and Perkins [89] note that the major difference between low-road and high-road transfer is that the latter involves mindful abstraction of the generic features of content, a chain of processing that is quite different from the spontaneous, automatic extension of learning, that they refer to as low-road transfer .

**2.2 Analysis of the ABR process**

Some believe that all intellectual acts involve analogical reasoning (e.g. see [[103]). Although this claim is open to debate, it is clear that much of our cognitive activity does depend on our ability to reason analogically. According to Mason [57] analogical reasoning helps learning in a relational way, i.e. connecting pieces of knowledge, when given reference systems provide means to penetrate and structure new domains.

Several studies (e.g. [23-24], [35], [62], [112], etc) have provided detailed models of the transfer process along these lines (*analogical transfer*). These models are broadly consistent with reviews of problem solving strategy training studies, in which factors associated with instances of successful transfer are identified. Summarizing the conclusions of the above studies one could state that the main stages involved in analogical transfer include:

- *Representation* of the target problem.
- *Search-retrieval* of the analogous problem.
- *Mapping* of the representations of the target and the possible analogous problem.
- *Adaptation* of the solution of the analogous problem for use with the target problem.

More specifically, before the solvers start working on the solution of a problem they usually construct a representation of it. A good representation must include both the surface and structural (abstract, solution relevant) features of the problem. The former are mainly determined by what are the quantities involved in the problem and the latter by how these quantities are related to each other.

As it is realized at least as early as 1945 by Duncker [19] this representation varies across solvers depending on their expertise with respect to the problem's domain. More recent studies (e.g. [37], [97], etc) supported Dunker's claims, showing that novices' representations primarily contain information about surface features of the problem, while experts' representations include also its structural features.

The features included in solvers' representations of the target problem are used as retrieval cues for a related problem in memory (called the *source or base problem*).

If a potential source problem is retrieved, solvers attempt to map the representations of the source and of the target problem in order to identify objects and relations that are in one-to-one correspondence.

Next, if the correspondences identified are such that the source problem can be considered as analogous to the target, solvers attempt to adapt the solution procedure of the source for use with the target problem .To determine the solution of the target by analogy to the source problem, the correspondences between objects and relations of the two problems must be used. The successful completion of this process is referred as *positive analogical transfer*.

But the search may also yield *distracting problems*, having surface but not structural common features with the target problem, and therefore being only superficially similar to it. Usually the reason for this is a non satisfactory representation of the target problem, containing only its salient surface features and the resulting consequences on the retrieval cues available for the search process.

When a distracting problem is considered as an analogue of the target, we speak about *negative analogical transfer*. This happens if a distracting problem is retrieved as a source problem and the solver fails, through the mapping of the representations of the source and target problem, to realize that the source cannot be considered as an analogue to the target problem. Therefore the process of mapping is very important in analogical problem solving, because it plays the role of a *control system* for the fitness of the source problem.

**2.3 Classroom experiments**

A series of experiments (e.g. [31], [36], [80], etc.) has proved with clarity that, when the source and target problems share both surface and structural features, spontaneous positive transfer should be expected regardless of expertise. In fact, although the subjects in these experiments were most likely novices, it seems reasonable to infer that, if they showed positive transfer under these favourable conditions, experts would have also.

Novick [62] claims further that, when the target and the source problem share structural, but not surface features (in this case the source, according to Holyoak's [35] terminology, is called a *remote analogue* of the target problem), spontaneous positive transfer should be more likely in experts than in novices. In contrast, when the source happens to be a distracting problem, spontaneous negative transfer should be stronger for novices than for experts. Novick [62] supports her claims by the results of three experiments, where the subjects were undergraduate students of the Universities of Stanford and Los Angeles.

In [113] we have also performed four classroom experiments with subjects undergraduate students of the School of Management and Economics of the Technological Educational Institute of Messolonghi, Greece. The results of these experiments gave a strong indication that the rendering of students by the teacher about the analogical problem solving process (steps of ABR, presentation of suitable examples, etc), improves significantly the novices', but not the experts' performance as well. This may be explained by the fact that probably most of the experts had already (before the teacher's rendering) assimilated empirically the analogical way of thinking. On the other hand, the fact that the teacher's rendering improves significantly the novices' performance, underlines its necessity, since the teacher must be addressed to all his (her) students.

Further, as it turned out from the statistical evaluation of the outcomes of the four experiments, the unsuccessful solvers encountered difficulties mainly at the step of mapping and less at the steps of representation of the target and of the adaptation of the solution of the analogous problem. This looks logical, because, as it turns out from the analysis (in section 2.2) of the ABR process, mapping is the most difficult step, since it requires an increased ability for abstraction from the solver.

## 3. Case – Based Reasoning

### 3.1 General characteristics

CBR is consistent with much that psychologist have observed in the natural problem solving that people do. People tend to be comfortable using CBR methodology for decision making, in dynamically changing situations and other situations were much is unknown and when solutions are not clear.

In CBR's terminology, a *case* denotes a problem situation. A previously experienced situation, which has been captured and learned in a way that it can be reused in the solving of future problems, is referred as a *past case , previous case, stored case, or retained case*. Correspondingly, a *new case, or unsolved case*, is the description of a new problem to be solved. The CBR systems' expertise is embodied in a collection (*library*) of past cases rather, than being encoded in classical rules. Each case typically contains a description of the problem plus a solution and/or the outcomes. The knowledge and reasoning process used by an expert to solve the problem is not recorded, but is implicit in the solution.

A lawyer, who advocates a particular outcome in a trial based on legal precedents, or an auto mechanic, who fixes an engine by recalling another car that exhibited similar symptoms, or even a physician, who considers the diagnosis and treatment of a previous patient having similar symptoms, to determine the disease and treatment for the patient in front of him, are using CBR; in other words CBR is a prominent kind of analogy making.

There are two styles of CBR; *problem solving style* and *interpretive style*. PS style can support a variety of tasks including planning, diagnosis and design (e.g. in Medicine [99], Industry [34] and Robotics [29]). The interpretive style is useful for (a) situation classification, (b) evaluation of solution, (c) argumentation, (d) justification of solution interpretation or plan and (e) the projection of effects of a decision of plan. Lawyers and managers making strategic decisions use the interpretive style ([79], [83]).

CBR is liked by many people, because they feel happier with examples rather, than conclusions separated from their context. A case-library can also be a powerful corporate resource allowing everyone in an organization to tap in the corporate library, when handling a new problem. CBR allows the case-library to be developed incrementally, while its maintenance is relatively easy and can be carried out by domain experts.

CBR is often used where experts find it hard to articulate their thought processes when solving problems. This is because knowledge acquisition for a classical knowledge-based system would be extremely difficult in such domains, and is likely to produce incomplete or inaccurate results. When using CBR the need for knowledge acquisition can be limited to establishing how to characterize cases. Some of the characteristics of a domain that indicate that a CBR approach might be suitable include: Records of previously solved problems exist, historical cases are viewed as an asset which ought to be preserved, remembering previous experiences is useful (experience is at least as valuable as textbook knowledge), specialists talk about the domain by giving examples.

CBR's coupling to *learning* occurs as a natural by-product of problem solving. When a problem is successfully solved, the experience is retained in order to solve similar problems in future. When an attempt to solve a problem fails, the reason for the failure is identified and remembered in order to avoid the same mistake in

Voskoglou & Salem

future. This process was termed as *failure-driven learning* [94]. Thus CBR is a cyclic and integrated process of solving a problem, learning from this experience, solving a new problem, etc. Effective learning in CBR, sometimes referred as *case-based learning*, requires a well worked out set of methods in order to extract relevant knowledge from the experience, integrate a case into an existing knowledge structure and index the case for later matching with similar cases.

The driving force behind case-based methods has to a large extent come from the *machine learning* community, and CBR is regarded as a subfield of machine learning. In fact, the notion of CBR does not only denote a particular reasoning method, irrespective of how the cases are acquired, it also denotes a machine learning paradigm that enables sustained learning by updating the case base after a problem has been solved.

## 3.2 History of CBR

The first trails into the CBR field have come from the study of analogical reasoning (see section 2) and –further back – from theories of concept formation, problem solving and learning within philosophy and psychology (e.g. [102], [123], etc). For example, Wittgenstein [123] observed that concepts, which are part of the natural world, like bird, tree, chair, car, etc, are polymorphic and therefore it is not possible to come up with a classical definition, but it is better to be defined by their sets of instances, or cases.

Memory is the repository of knowledge and therefore the question is what kind of memory accounts for observed cognitive behaviors. A leading theory has been the *semantic memory* model. Psychologists devoted much attention to this theory ([17], [42], [82], etc), as have Artificial Intelligence (AI) researchers ([73], [124], etc), who attempted to create computer programs that model cognitive processes. The semantic memory model typically represents static facts about the world and therefore this type of knowledge does not change over time. However it was observed that this model did not account for all the data; e. g. it does not explain how knowledge is incorporated into memory and where does the information come from.

To address these and other questions Tulvin [108-109] proposed a theory of *episodic memory* as an adjunct to semantic memory. Episodic memory receives and stores information about temporally dated episodes or events. The retrieval of information from the episodic store serves as a special type of input into episodic memory and thus changes the contents of the episodic memory store.

CBR traces its roots in *Artificial Intelligence* (AI) to the work of Roger Schank and his students at Yale University – U.S.A. in the early 1980's. Schank [90] proposed a *conceptual memory* that combined semantic memory with Tulvin's episodic memory. *Scripts* [91] were proposed as a knowledge structure for the conceptual memory. The acquisition of scripts, which are analogous to Minsky's [59] *frames*, is the result of repeated exposure to a given situation. As a psychological theory of memory scripts suggested that people would remember an event in terms of its associated script. However an experiment by Bower et al. [14] showed that subjects often confused events that have similar scripts: e. g. one might mix up waiting room scenes from a visit to a doctor with a visit to a dentist. These data required a revision in script theory. Schank [92-93] postulated a more general structure to account for the diverse and heterogeneous nature of episodic memory, called *memory organization packet (MOP)*. MOP's can be viewed as meta-scripts; e. g. a professional office visit MOP can be instantiated and specified for both the doctor and the dentist, thus providing the basis for confusion between these two events.

However, more important than the MOP knowledge was the new emphasis on the basic memory processes of reminding and learning. Schank proposed a theory of learning based on reminding, according to which we can classify a new episode in terms of past similar cases. Schank's model of *dynamic memory* [95] was the basis of the earliest CBR systems that might be called *case-based reasoners*: Kolodner's CYRUS [45] and Lebowitz's IPP [50]. The basic idea of Schank's model [95] is to organize specific cases, which share similar properties, under a more general structure called a *generalized episode (GE)*. During storing of a new case, when a feature of it matches a feature of an existing past case, a new GE is created. Thus the organization and structure of memory is dynamic, i. e. changes over time. Similar parts of two case descriptions are generalized in to a new GE and the cases are indexed under this GE by their different features. Concerning CYRUS, it was basically a question-answering system with knowledge of the various travels and meetings of former US Secretary of State Cyrus Vance and the case memory model developed for this system has later served as basis for several other CBR systems including MEDIATOR, PERSUADER, JULIA,, etc.

An alternative approach for the representation of cases in a CBR system is the *category and exemplar model,* produced by the work of Bruce Porter and his group at the University of Texas. In this model the case memory is embedded in a network of categories, cases and index pointers. Each case is associated with a category. Finding a case in the case library that matches an input description is done by combining the features of the new problem case into a pointer to the category that shares most of these features. A new case is stored in a category by searching for a matching case and by establishing the appropriate feature indices. The above model applied first to the PROTOS system ([13], [72]), where emphasis is given to the combination of the general with the specific knowledge obtained through the study of cases.

 Another case memory model was produced by the work of Edwina Rissland and her group at the University of Massachusetts, interested in the role of precedence reasoning in legal judgments [78]. This work resulted in the HYPO [10] and CABARET [100] systems, where cases are grouped under a set of domain-specific dimensions.

Other early significant contributions to CBR include,  the *Memory-Based Reasoning (MBR)* model of Stanfill and Waltz [104], designed for parallel computation rather than knowledge-based matching, the study of Phyllis Koton at MIT on the use of CBR to optimize performance in an existing knowledge based system resulted in the CASEY system [49], etc.

In Europe research on CBR was taken up a little later, to a large extend focused towards the utilization of knowledge level modeling in CBR systems. Among the earliest results was the work of Althoff , Richter and others at the University of Kaiserslautern for complex technical diagnosis within the MOLTKE system [8], which lead to the PATDEX system [75], and later to several other systems and methods. In Blanes, Plaza and Lopez developed a learning apprentice system for medical diagnosis [65], while in Aberdeen Sleeman's group studied the use of cases for knowledge base refinement (REFINER system [98]).

At the University of Trondheim Aamodt and colleagues at Sintef studied the learning aspect of CBR in the context of knowledge acquisition and maintenance, while for PS the combined use of cases and general domain knowledge was focused [1] This lead to the development of CREEK system and to continued work on knowledge-intensive CBR. On the cognitive science side significant work was done on analogical reasoning at Trinity College, Dublin [38] and by Strube's group at the University of Freiburg, where the role of episodic knowledge in cognitive models was investigated in the EVENTS project [107].

Currently, the CBR activities in the USA as well as in Europe are spreading out and the number of papers on CBR in almost any AI journal is rapidly growing. Germany seems to have taken a leading position in terms of active researchers and several research groups of significant activity level have been established recently. The basic ideas and the underlined theories of CBR have spread quickly to other continents as well; from Japan, India [111] and other Asian countries, there are also activity points. In Japan the interest is mainly focused towards the parallel computation approach in CBR [43].

In the 1990's , interest in CBR grew in the international community, as evidenced by the establishment of an International Conference on CBR in 1995, as well as European, German, British, Italian and other CBR workshops.

We must mention also the existence of a continuously increasing number of *websites* that include many references and links to electronic CBR resources, such us the US Navy Research Website, the University of Kaiserslautern Website, the AI-CBR Website of the University of Salford, including a mailing list with announcements, questions and discussions about CBR, the CBR Newsletter, that originated as a publication of the Special Interest Group on CBR in the German Society for Computer Science, the Web server of the CBR Group, part of the Department of Computer Science at the University of Massachusetts at Amherst, the Website of the AI Applications Institute (AIAI), part of the School of Informatics at the University of Edinburgh, the official server of the International Conference of CBR, the AI-CBR Website of the department of Computer Science at the University of Auckland, the Machine Learning Network on line Information Service, etc . Some of the above websites are listed in detail in our references section ([7], [9], [16], [54], [63]).

## 3.3 The steps of the CBR process

CBR has been formalized for purposes of computer and human reasoning as a four step process, known as the *dynamic model of the CBR cycle*. These steps involve the following actions:

- *Retrieve* the most similar to the new problem past case, or cases.
- *Reuse* the information and knowledge in that case to solve the problem.
- *Revise* the proposed solution.
- *Retain* the parts of this experience likely to be useful for future problem-solving.

In more detail, an initial description of a problem defines a new case. This new case is used to *retrieve* the most similar case, or cases, from the library of previous cases. The subtasks of the retrieving procedure involve: Identifying a set of relevant problem descriptors, matching the case and returning a set of sufficiently similar cases, given a similarity threshold of some kind, and selecting the best case from the set of cases returned. Some systems retrieve cases based largely on superficial syntactic similarities among problem descriptors, while advanced systems use semantic similarities.

The retrieved case (or cases) is combined, through *reuse*, with the new case into a solved case, i.e. a proposed solution of the initial problem. The reusing procedure focuses on identifying the differences between the retrieved and the current case, as well as the part of the retrieved case which can be transferred to the new case. CBR methods are implemented by retrieval methods (to retrieve past cases), a language of preferences (to select the best case) and a form of derivational analogy (to reuse the retrieved method into the current problem).

Through the *revise* process this solution is tested for success, e.g. by being applied to the real world environment, or a simulation of it, or evaluated by a teacher, and repaired, if failed. This provides an opportunity to learn from failure.

During the *retain* action useful experience is retained for future reuse, and the case base is updated by a new learned case, or by modification of some existing cases. The retaining process involves deciding what information to retain and in what form to retain it, how to index the case for future retrieval, ant integrating the new case into the case library.

The *general knowledge* usually plays a part in the CBR cycle by supporting the CBR process. This support however may range from very weak (or none) to very strong, depending on the type of the CBR method. By general knowledge we here mean general, domain-dependent knowledge, as opposed to specific knowledge embodied by cases. For example, in the case a lawyer, mentioned in our introduction, who advocates a particular outcome in a trial based on legal precedents, the general knowledge is expressed through the existing relevant laws and the correlations between them and the case of the trial. A set of rules may have the same role in other CBR cases.

While the process-oriented view of CBR presented above enables a global, external view to what is happening, a task-oriented view could be suitable for describing the detailed mechanisms from the perspective of the CBR reasoner itself. This is coherent with the task-oriented view of knowledge level modeling, where a system is viewed as an agent which has goals, and means to achieve its goals. Tasks are set up by the goals of the system and a task is performed by applying one or more methods.

Such a *task-method decomposition* of the four main steps of the CBR process to sub-steps, where related problem-solving methods are also described, is given – in the form of a decision tree - in Aamodt & Plaza ( [3]; Figure 2). The top-level task is *problem-solving* and *learning from experience* and the *method* to accomplish the task is CBR. This splits the top-level task into the four major CBR tasks: *retrieve, reuse, revise and retain*. All the four tasks are necessary in order to perform the top-level task. The retrieve task is, in turn, partitioned into the subtasks *identify features* (collect descriptors, interpret problem, infer descriptors), *search* (to find a set of past cases), *initially match* (calculate and/or explain similarity), and *select* (the most similar case). In the same manner the reuse task is partitioned into the subtasks *copy* and *adapt* (the solution of the most similar case), the revise task is partitioned into the subtasks *evaluate solution* and *repair fault*, and the retain task is partitioned into the subtasks *integrate* (rerun problem, update general knowledge, adjust indexes), *index* (generalize and determine indexes) and *extract* (relevant descriptors, solutions, justifications and solution method). All task partitions are complete, i.e. the set of subtasks is intended to be sufficient to accomplish the task.

A *method* specifies the algorithm that identifies and controls the execution of subtasks, and accesses and utilizes the knowledge and information needed to do this. The methods shown in Aamodt's and Plaza's scheme [3], which are task decomposition and control methods, are actually high level method classes, from which one or more methods should be chosen. In this sense the method set, as shown in the scheme, is incomplete, i.e. one of the methods indicated may be sufficient to solve the task in a certain particular case, several methods may be combined, or there may be other methods that can do the job. For example, for the subtask "evaluate solution" of the task "revise" the evaluation could be done, according to the current problem, either by the teacher, or in real world, or/and in model. Another possible method, which is not shown into the scheme, is to evaluate the solution through simulation. In the same way, for the subtask "repair fault" this could be a self-repair, or a user-repair, etc.

A spherical observation of the task-oriented view of CBR described above, as Aamodt and Plaza [3] themselves accept, makes evident that their framework and analysis approach is strongly influenced by knowledge level modeling methods in general and by the Components of Expertise methodology in particular [105-106]. The following functional diagram of Figure 1 (where boxes represent process and ovals represent knowledge sources)., adapted from [47-48] and presented by Prof. Salem in his plenary lecture [88] at the 16$^{th}$ WSEAS International Conference on Computers (Kos island, Greece, July 14-17, 2012) gives a graphical representation of the CBR methodology: When a new problem is introduced in the system, the problem is indexed, and subsequently, the indexes are used to retrieve past cases from memory. These past cases lead to a set of prior solutions. Subsequently, the previous solutions are modified to adapt to the new situation. Then the proposed solution is tried out. If the solution succeeds, then it is stored as a working solution; if it fails, the working solution must be repaired and tested again.

In support of CBR processes, the following knowledge structures are necessary:

1. *Indexing Rules Knowledge Structure (IRKS)*: Indexing rules identifies the predictive features in the input that provides appropriate indexes into the case memory.

2. *Case Memory Knowledge Structure (CMKS)*: Case memory is the episodic memory, which comprises of the database of experience.

3. *Similarity Rules Knowledge Structure (SRKS)*: If more than one case is retrieved from episodic memory, the similarity rules (SR) can be used to decide which case is more like the current situation. For example: In the air shuttle case, we might be reminded of both airplane rides and train rides. The SR might initially suggest that we rely on the air plane case.

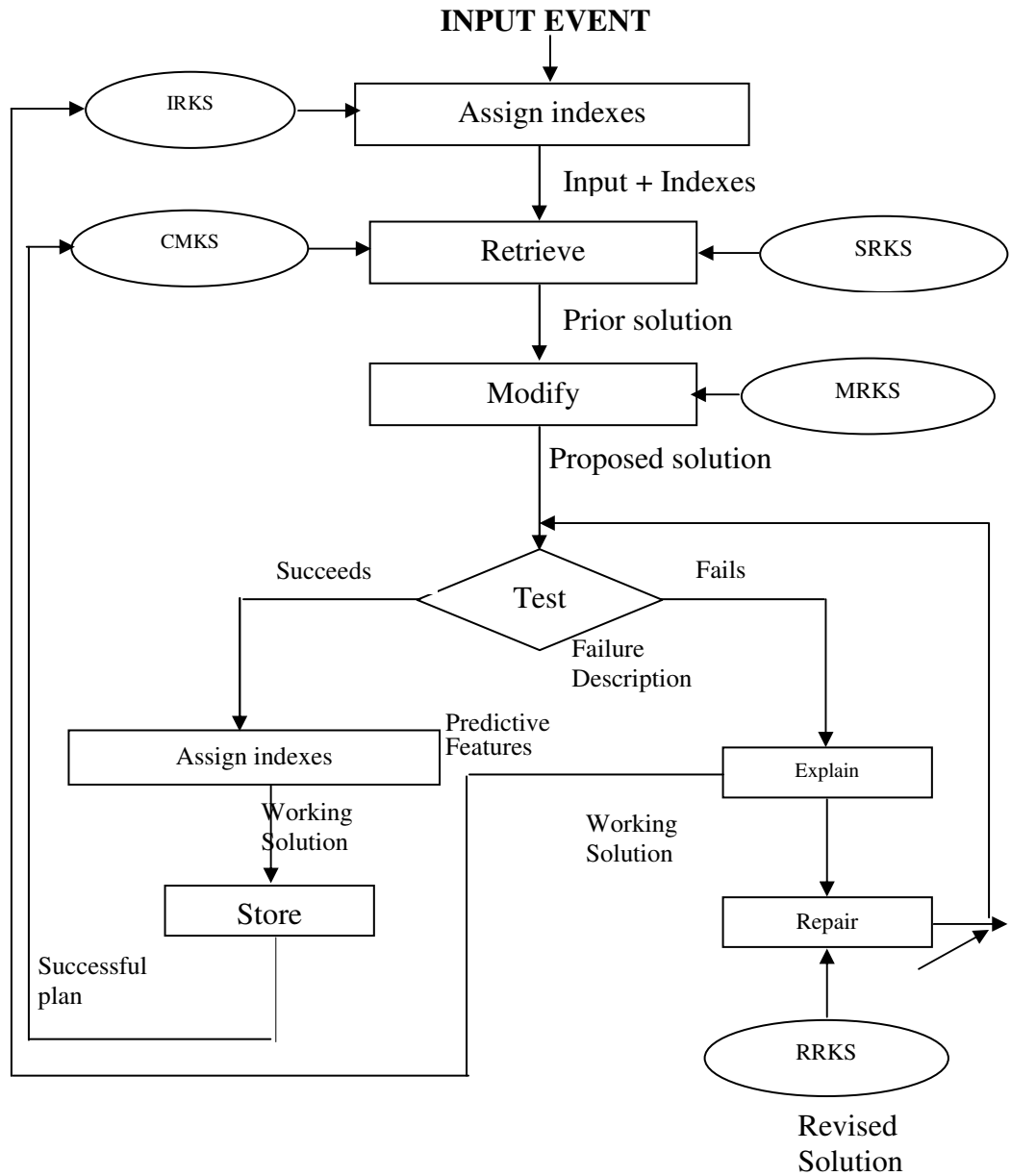

**Fig. 1: A functional diagram of the CBR methodology**

4. *Modification Rule Knowledge Structure (MRKS)*: If No "old case" is going to be an exact match for a new situation, the "old case" must be modified to fit. We require knowledge about what kinds of factors can be changed and how to change them. For the airplane example: It is acceptable to ride in a different seat, but it is usually not advisable to change roles from passenger to pilot.

5. *Repair Rules Knowledge Structure (RRKS)*: Once we identify and explain an expectation failure, we must try to alter our plan to fit the new situation. Again we have rules for what kinds of changes are permissible.

Other flowcharts illustrating the basic steps of the CBR process were produced by Riesbeck and Bain [76], Slade [101], Lei et al. [51], Voskoglou [116], etc.

**3.4 Main types of CBR methods**

In line with the descriptive framework for CBR presented above, core problems addressed by CBR research can be grouped into five areas: Knowledge representation, retrieval methods, reuse methods, revise methods ant retain methods. In a book published by Janet Kolodner [48], a member of Schank's research team, these problems are discussed and elaborated to substantial depth, and hints and guidelines on how to deal with them are given. An overview of the main problem issues related to these five areas is also given in Aamond & Plaza [3] with illustrating examples drawn from the systems PROTOS, CHEF, CASEY, PATDEX, BOLERO and CREEK.

A set of coherent solutions to these problems constitutes a *CBR method*.

As for AI in general, there are no universal CBR methods for every domain of application. The challenge in CBR is to come up with methods that are suited for problem-solving and learning in particular subject domains and for particular application environments. Thus the CBR paradigm covers a range of different methods for organizing, retrieving, utilizing and indexing the knowledge in past cases. Actually CBR is a term used both as a generic term for the several types of these methods, as well as for one such type described below, and this has lead to some confusion. Throughout this paper we are using the term CBR in the generic sense.

The main types of CBR methods are listed below:

- *Case-Based Reasoning*.

The typical CBR methods have three characteristics that distinguish them from the other approaches listed below. First, it is assumed to have a *complexity* with respect to their internal organization, i.e. a feature vector holding some values and a corresponding class is not what we would call a typical CBR description. Second, they are able to *modify*, or adapt a retrieved solution when applied in a different problem-solving context, and third they utilize *general background knowledge*, although its richness and role within the CBR processes vary. Core methods of typical CBR systems borrow a lot from cognitive psychology theories.

- *Analogy-Based Reasoning*

See section 2.

- *Exemplar-Based Reasoning.*

In the exemplar view a concept is defined extensionally as the set of its exemplars. In this approach solving a problem is a *classification task*, i.e. finding the right class for the unclassified exemplar. The set of classes constitutes the set of *possible solutions* and the class of the most similar past case becomes the solution to the classification problem. Modification of a solution found is therefore outside the scope of this method. Characteristic examples are the paper by Kibler and Aha [41], and the book of Bareiss [13].

- *Instance-Based Reasoning.*

This is a specialization of exemplar-based reasoning. To compensate for lack of guidance from general background knowledge, a relatively large number of instances is needed in order to close in on a concept definition. The representation of the instances is usually simple (e.g. feature vectors), since a major focus is to study *automated learning,* with no user in the loop. An example is the work by Aha et al. [6], and serves to distinguish their methods from more intensive exemplar-based approaches.

- *Memory-Based Reasoning.*

This approach emphasizes a collection of cases as a *large memory*, and reasoning as a process of accessing and searching in this memory. The utilization of *parallel processing* techniques is a characteristic of these methods and distinguishes this approach from the others ([43], [46], [104], etc). The *Massive Memory*

*Architecture* [66] is an integrated architecture for learning and PS based on reuse of case experiences retained in the systems memory. A goal of this architecture is the understanding and implementing the relationship between learning and PS into a reflective or introspective framework: the system is able to inspect its own past behavior in order to learn how to change its structure so as to improve its future performance.

Most CBR systems make use of general domain knowledge in addition to knowledge represented by cases. Representation and use of that domain knowledge involves *integration* of the case-based method with other methods and representation of problem-solving, for instance rule-based systems or deep models like casual reasoning. The overall architecture of the CBR system has to determine the interactions and control regime between the CBR method and the other components.

For instance, the CASEY system integrates a model-based causal reasoning program to diagnose heart diseases. When the case-based method fails to provide a correct solution, CASEY executes the model-based method to solve the problem and stores the solution as a new case for future use. Another example of integrating rules and cases is the BOLERO system [53], which has a meta-level architecture, where the base-level is composed of rules embodying knowledge to diagnose the plausible pneumonias of a patient, while the meta-level is a case-based planner that, at every moment is able to dictate which diagnoses are worthwhile to consider. In the CREEK architecture, the cases, heuristic rules, and deep models are integrated into a unified knowledge structure. The main role of the general knowledge is to provide explanatory support to the case-based processes [2]; rules or deep models may also be used to solve problems on their own, if the case-based method fails. This line of work has also being developed in Europe by systems like the Massive Memory Architecture and INRECA [55]. In these systems, which are closely related to the multi-strategy learning systems [58], the issues of integrating different PS and learning methods are essential.

## 3.5 Tools and applications of CBR

A CBR tool should support the four main processes of CBR: retrieval, reuse, revision and retention. A good tool should support a variety of retrieval mechanisms and allow them to be mixed when necessary. In addition, the tool should be able to handle large case libraries with the retrieval time increasing linearly (at worst) with the number of cases. CBR first appeared in commercial tools in the early 1990's and since then has been sued to create numerous applications in a wide range of domains. Organizations as diverse as IBM, VISA International, Volkswagen, British Airways and NASA have already made use of CBR in applications such as customer support, quality assurance, aircraft maintenance, process planning and decision support, and many more applications are easily imaginable. At Lokheed, Palo Alto, a fielded CBR system was developed. The problem domain is optimization of autoclave loading for heat treatment of composite materials [33]. The autoclave is a large convection oven, where airplane parts are treated in order to get the right properties. Different material types need different heating and the task is to select the parts that can be treated together and distribute them into the oven so that their required heating profiles are taking care of. A second fielded CBR system has been developed at General Dynamics, Electric Boat Division [15] handling the problem of the selection of the most appropriate mechanical equipment during the construction of ships, and to fit it to its use. Most of these problems can be handled by fairly standard procedures, but some of them, referred as "non-conformances", are harder and occur less frequently. In the period December 1990 – September 1991 20000 non-conformances were handled through the prototype CBR system that was developed and the cost reduction, compared to previous costs of manual procedures, was about 10%, which amounts to a saving of $240000 in less than one year.

In general the main domains of the CBR applications include *diagnosis, help-desk, assessment, decision support, design*, etc.

More explicitly:

CBR diagnostic systems try to retrieve past cases, whose symptom lists are similar in nature to that of the new case and suggest diagnoses based on the best matching retrieved cases. CBR diagnostic systems are also used in the customer service area dealing with handling problems with a product or service (help-desk applications), e.g. Compaq SMART system [61].

In the assessment processes CBR systems are used to determine values for variables on comparing it to the known value of something similar. Assessment tasks are quite common in the finance and marketing domains.

In decision making, when faced with a complex problem, people often look for analogous problems for possible solutions. CBR systems have been developed to supporting this problem retrieval process to find relevant similar problems. CBR is particularly good at querying structured, modular and non-homogeneous documents. A number of CBR decision support tools are commercially available, including k-Commerce from eGam, Kaidara Advisor from Kaidara and SMART from Illation.

Finally, systems to support human designers in architectural and industrial design have been developed. These systems assist the user in only one part of the design process, that of retrieving past cases, and would need to be combined with other forms of reasoning to support the full design process. An early such example is Lockheed's CLAVIER, a system for laying out composite parts to be baked in an industrial convection oven [56].

Several commercial companies offer *shells* for building CBR systems. Just as for rule-based systems shells, they enable you to quickly develop applications, but at the expense of flexibility of representation, reasoning approach and learning methods. Four such shells are reviewed in [30]: ReMind from Cognitive Systems Inc., CBR Express/ART-IM from Inference Corporation, Esteem from Esteem Software Inc., and Induce-it (later renamed to Case-Power) from Inductive Solutions Inc. On the European scene Acknosoft in Paris offers the shell KATE-CBR as part of their Case-Craft Toolbox, Isoft, also in Paris, has a shell called

ReCall, TecchInno in Kaiserslauten has S3-Case, a PATDEX-derived tool that is part of their S3 environment for technical systems maintenance.

Some academic CBR tools are freely available, e.g. the PROTOS system [72], which emphasized on integrating general domain knowledge and specific case knowledge into a unified representation structure, is available from the University of Texas, and code for implementing a simple version of dynamic memory, as described in [77], is available from the Institute of Learning Sciences at Northwestern University.

A book has been published by Ian Watson [122] in which the author explains the principles of CBR by describing its origins and constructing it with familiar information disciplines such as traditional data processing, logic programming, rule-based expert systems, and object-oriented programming. Through case studies and step-by-step examples, he goes on to show how to design and implement a reliable, robust CBR system in a real-world environment. Additional resources are provided in a survey of commercially available CBR tools, a comprehensive bibliography, and a listing of companies providing CBR software and services.

## 4. Applications of CBR in Medical Domain

CBR has already been applied in a number of different applications in medicine. Some CBR systems used in medical applications are: CASEY that gives a diagnosis for the heart disorders [48], GS.52 which is a diagnostic support system for dysmorphic syndromes, NIMON which is a renal function monitoring system, COSYL that gives a consultation for a liver transplanted patient [52], ICONS that presents a suitable calculated antibiotics therapy advise for intensive care patients [32], etc. In the next two sub-sections we present briefly two cases of CBR systems developed in the Ain Shams University, Egypt for medical applications

**4.1 CBR-based system for diagnosis of cancer diseases**

Cancer is a group of more than 200 different diseases; it occurs when cells become abnormal and keep dividing and forming either benign or malignant tumors. Cancer has initial signs or symptoms if any is observed, the patient should perform complete blood count and other clinical examinations. Then to specify cancer type, patient needs to perform special lab-tests.

This section presents a summary of the CBR-based expert system prototype for diagnosis of cancer diseases developed by Bio-Medical Informatics and Knowledge Engineering Labs at Artificial Intelligence Research Unit, Faculty of Computer and Information Sciences, Ain Shams University, Cairo, Egypt. The main purpose of the system is to serve as doctor diagnostic assistant. The system provides recommendation for controlling pain and providing symptom relief in advanced cancer. It can be used as a tool to aid and hopefully improve the quality of care given for those suffering intractable pain. The system is very useful in the management of the problem, and its task to aid the young physicians to check their diagnosis ([84], [86]).

Figure 2 shows the architecture of the CBR-based system for cancer diagnosis. The system's knowledge base is diverse and linked through a number of indices, frames and relationships. The bulk of this knowledge consists of actual case histories and includes 70 cancer patient cases; some are real Egyptian cases and some from virtual hospitals on the internet. The system consists of three main modules; user interface, case base reasoning module and computational module; all are interacted with the main environment of cancer diseases. The user is a cancer expert doctor, the interaction is through menus and dialogues that simulate the patient text sheet containing symptoms and lab examinations. Computational model uses rule-based inference to give diagnostic decision and each new case is stored in case library. Patient cases are retrieved in dialogue with similarity matches using the nearest neighbor matching technique. The initial diagnostic process is done through firing of rules in the Rule-Based inference. These rules encode information about patient's symptoms and pathological examinations. Frames technique is used [120] for patient case indexing, storage and retrieval.

Voskoglou & Salem

The patient case will include age, sex and weight occupation, pathologic, medical history family, physical exams and treatments.

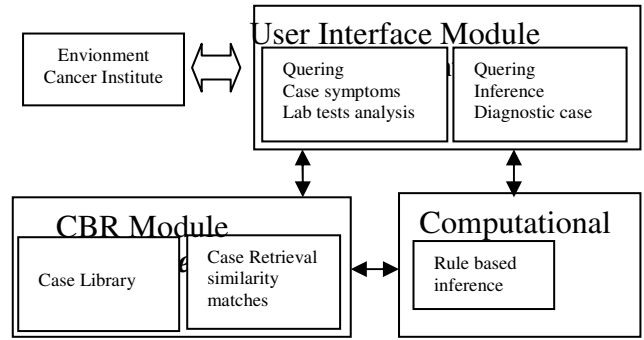

**Fig.2: Architecture of the CBR-based system for cancer diagnosis**

Below a typical example is given .of an Egyptian liver cancer case description of an old woman:

> Patient: 65-years old female not working, with nausea and vomiting.
>
> Medical History: Cancer head of pancreas
>
> Physical Exam: Tender hepatomgaly liver, large amount of inflammatory about 3 liters, multiple liver pyogenic abscesses and large pancreatic head mass.
>
> Laboratory Findings: Total bilrubin 1.3 mg/dl, direct bilrubin 0.4 mg/dl, sgot (ast) 28 IU/L, sgpt (alt) 26 IU/L.

**4.2 CBR-based system for diagnosis of heart diseases**

Heart disease is a vital health care problem affecting millions of people. Heart disease are of 25 different ones; e.g. left-sided heart failure, right-sided heart failure, angina pectoris, myocardial infraction and essential hypertension. The system is able to give an appropriate diagnosis for the presented symptoms, signs and investigations done to a cardiac patient with the corresponding certainty factor. It can be used to serve as doctor diagnostic assistant and support the education for the undergraduate and postgraduate young physicians.

In this system the knowledge is represented in the form of frames and the case memory contains 110 cases for 4 heart diseases namely; mistral stenosis, left-sided heart failure, stable angina pectoris and essential hypertension. Each case contains 207 attributes concerning both demographic and clinical data. After removing the duplicate cases, the system has trained set of 42 cases for Egyptian cardiac patients. Statistical analysis has been done to determine the importance values of the case features. Two retrieval strategies were investigated namely; induction and nearest neighbor approaches. The results indicate that the nearest neighbor is better than the induction strategy. Cardiologists have evaluated the overall system performance where the system was able to give a correct diagnosis for thirteen new cases [85].

**4.3 Knowledge Engineering Issues in Developing Biomedical CBR Systems**

*Knowledge engineering* (KE) was defined in 1983 by Feigenbaum and McCorduck [20] as follows: KE is an engineering discipline that involves integrating knowledge into computer systems in order to solve complex problems normally requiring a high level of human expertise

It follows a brief discussion of the knowledge engineering issues which are crucial in developing CBR Systems for any healthcare task [4-5].

1. *Case Representation:* Determining the appropriate case features is the main knowledge engineering process in CBRS. The case is a list of features that lead to a particular outcome (e.g. the information on a patient history and the associated diagnosis). This process involves; (a) defining the terminology of the domain and (b) gathering representative examples of problem solving by the expert. Representations of cases can be in any of several forms; predicate representations, frame representations and representations resembling database entries.
2. *Case Indexing Process:* The CBRS derives its power from its ability to retrieve relevant cases quickly and accurately from its memory. Figuring out when a case should be selected for retrieval in similar future situations is the goal of the case *indexing process*. Building a structure or process that will return the most appropriate case (from the case memory) is the goal of the *retrieval process*. Case indexing process usually falls into one of three approaches: nearest neighbor, inductive, and knowledge-guided or a combination of the three.

3. *Case Memory Organization and Retrieval:* Once cases are represented and indexed, they can be organized into an efficient structure for retrieval. Most case memory structures fall into a range between *purely associative retrieval*, where any or all of the features of a case are indexed independently of the other features and *purely hierarchical retrieval*, where case features are highly organized into a general-to-specific a concept structure. Nearest-neighbor matching techniques are considered associative because they have no real-memory organization. Discrimination nets are more of a cross between associative and hierarchical because they have some structure to the net but greater retrieval flexibility because they have a greater number of links between potential indexing features. Decision trees are an example of purely hierarchical memory organization.

4. *Case Adaptation:* It is difficult to define a single generically applicable approach to perform case adaptation, because adaptation tends to be problem specific. Most existing CBR systems achieve case adaptation for the specific problem domains they address by encoding adaptation knowledge in the form of a set of adaptation rules or domain model. Adaptation rules are then applied to a retrieved case to transform it into a new case that meets all of the input problem's constraints. More recent applications have successfully used pieces of existing cases in memory to perform adaptations.

5. *Learning and Generalization:* As cases accumulate, case generalization can be used to define prototypical cases that embody the major features of a group of specific cases, and those prototypical cases can be stored with the specific cases, improving the accuracy of the system in the long run.

6. *CBR - Tools and Shells:* The availability of a commercial CBR shells in the market helps the knowledge engineers to overcome some of the problems they currently face in designing and maintaining large knowledge-base learning systems using rule based tools ([74], [81]).

**4.4 Benefits of the Expert Support Systems to Healthcare**

The benefits of using expert support systems approach in the healthcare sector are linked mainly with patients' treatment. We may enumerate several areas in which expert systems bring benefits, these are:

a)  *Treatment choice* – may be easier with the use of if-then rules of an expert system; Following the rules, a physician is able to infer treatment adequate to symptoms and/or to a specific illness;

b)  *Diagnosis support* – this comes both from rule-based systems as well from case based ones. If-then rules enable encoding of knowledge linking symptoms to illnesses, while case-based reasoning enables finding the illness by comparing patients' symptoms to these stored in case-based knowledge base;

c)  *Analysis of treatment options* – rule-based knowledge enables a so-called what-if analysis: what is probable to happen if we use a specific treatment?

d)  *Keeping medical history* – is easy with case-based expert systems, where individual patients' cases may be stored both for statistical purposes and for case-based reasoning.

**4.5 Conclusions**

CBR is an appropriate methodology for all medical domains and tasks for the following reasons: cognitive adequateness, explicit experience, duality of objective and subjective knowledge, automatic acquisition of subjective

Voskoglou & Salem

knowledge, and system integration. CBR presents an essential technology of building intelligent CBR systems for medical diagnoses that can aid significantly in improving the decision making of the physicians. These systems help physicians and doctors to check, analyze and repair their solutions. The physician inputs a description of the domain situation and his (her) solution and the system can recalls cases with similar solutions and presents their outcomes to the student. Also he (she) attempts to analyze the outcomes to provide an accounting of why the proposed type of solution succeeded or failed.

## 5. Development trends of CBR methods and applications

The development trends of CBR methods can be grouped around five main topics.

- *Integration with other learning methods* is the first topic that forms part of the current trend in research towards multi-strategy learning systems. This research aims at achieving an integration of different learning methods into a coherent framework, where each learning method fulfills a specific and distinct role in the system, e.g. case-based learning and induction as is done in MMA and INRECA systems.
- *Integration with other reasoning components* is the second topic that aims at using the different sources of knowledge in a more thorough, principal way, like what is done in the CASEY system with the use of causal knowledge. This trend, which is very popular in the European continent, emphasizes the increasing importance of knowledge acquisition issues and techniques in the development of knowledge-intensive CBR systems.
- The *massive memory parallelism* trend applies CBR to domains suitable for shallow, instance-based retrieval methods on a very large amount of data. This direction may also benefit from integration with neural network methods, as several Japanese projects currently are investigated [43].

- By the fourth trend, *method advances by focusing on the cognitive aspects*, in particular in the follow-up work initiated on creativity (e.g. [96]) as a new focus for CBR methods. It is not just an "application type", but a way to view CBR in general, which may have significant impacts on the CBR methods in future.

Finally, concerning the fifth topic one must notice that as a general PS methodology intended to cover a wide range of real-world applications, CBR must face the challenge to deal with uncertain, incomplete and vague information. In fact, successfully deployed CBR systems are commonly integrated with some method to treat uncertainty, which is already inherent in the basic CBR hypothesis demanding that similar problems have similar solutions. Correspondingly, recent years have witnessed an increased interest in formalizing parts of the CBR methodology within different frameworks of reasoning under uncertainty, and in building hybrid approaches by combining CBR with methods of uncertain and approximate reasoning.

- *Fuzzy logic* can be mentioned as a particularly interesting example. In fact, even though both CBR and fuzzy systems are intended as cognitively more plausible approaches to reasoning and problem-solving, the two corresponding fields have emphasized different aspects that complement each other in a reasonable way. Thus fuzzy set-based concepts and methods can support the various aspects of CBR including: Case and knowledge representation, acquisition and modeling, maintenance and management of CBR systems, case indexing and retrieval, similarity assessment and adaptation, instance-based and case-based learning, solution explanation and confidence, and representation of context. On the other way round ideas and techniques for CBR can contribute to fuzzy set-based approximate reasoning.

Notice that, in recent papers ([117], [119]) we have constructed fuzzy models for a more effective description of the ABR and CBR processes, in which their main steps (see sections 2.2 and 3.3 respectively) are represented as fuzzy subsets of a set of linguistic labels characterizing the success in each of these steps. Thus, by calculating the possibilities of all profiles, one can obtain a qualitative view of the evolution of the ABR /CBR process respectively. In the same papers we have also applied principles of fuzzy logic and of uncertainty theory (see the book [44]) in obtaining several measures for the effectiveness of the corresponding group of analogical problem-solvers or of the corresponding CBR system.

Notice also that in earlier papers ([112], [118]) we have constructed stochastic models for the description of the ABR and CBR processes. Namely, in each case we introduced a finite Markov chain having as states the steps of the corresponding process and, by applying basic principles of the relevant theory (e.g. see the book [40]), we succeeded in obtaining some quantitative results (probabilities etc) characterizing the ABR/CBR process respectively. However, and in contrast to the fuzzy models who give also a qualitative/realistic view of the corresponding situation, our stochastic models are self-restricted in giving quantitative information only about the 'ideal behavior' of the corresponding group/CBR system respectively. This is in general one of the main advantages of fuzzy logic with respect to probability theory in dealing with problems characterized by a degree of vagueness and/or uncertainty.

The trends of CBR applications clearly indicate that we will initially see a lot of help-desk applications around and this type of systems may open up for a more general coupling of CBR to information systems. The use of cases for human browsing and decision making is also likely to lead to an increased interest in intelligent computer-aided learning, training and teaching, since CBR systems are able to continually learn from and evolve through the capturing and retaining of past experiences. On the other hand, the diagnostic systems (mainly for medical purposes) and the legacy databases will continue to be some of the most common applications of CBR; for example AIAI at the School of Informatics of the University of Edinburgh [9], has successfully applied CBR to otherwise intractable problems such as fraud screening.

## 6. Final conclusions and discussion

In the present paper we reviewed ABR and CBR, two similar PS methods in which the key idea is to tackle new problems by referring to similar problems that have already been solved in the past.

The major focus of study in ABR has been on the reuse of a past case, what is called the mapping problem: Finding a way to transfer, or map, the solution of an identified analogue (source, or base problem)*,* to the present problem (target problem*)*. The main steps of the ABR process include representation of the target problem, search-retrieval for a related problem in memory, mapping of the common features of the source and of the target problem and adaptation of the solution procedure of the source problem for use with the target problem

Longstanding research in AI and related fields has produced a number of paradigms for building intelligent and knowledge-based systems, such as rule-based reasoning, constraint processing, or probabilistic graphical models. Being one of these paradigms CBR has received a great deal of attention in recent years and has been used successfully in diverse application areas. CBR proceeds from individual experiences in the form of cases. The generalization beyond these experiences is largely founded on principles of ABR in which the cognitive concept of similarity plays an essential role. CBR emphasizes PS and learning as two sides of the same process: PS uses the results of past learning episodes, while it provides the backbone of the experience from which learning advances.

The advantages and benefits of the CBR methodology can be summarized as follows:

1. Can make use of background domain knowledge when available.

2. It integrates symbolic and numeric techniques.

3. It supports fuzzy quantities and queries.

4. It offers rich indexing support.

5. It uses known solutions to past experiences for solving a new problem whose solution is unknown.

Voskoglou & Salem

6. It combines the benefits of information retrieval and rule based reasoning.

7. It copes with complex structured data

The current state of art in Europe regarding CBR is characterized by a strong influence of the USA ideas and CBR systems, although Europe is catching up and provides a somewhat different approach to CBR, particularly in its many activities related to integration of CBR and other approaches and by its movement toward the development of application-oriented CBR systems. The basic ideas of CBR have spread quickly to other continents; from Japan, India and other Asian countries there are also activity points.

The key difference between *CBR* and the classical *rule-induction algorithms*, which are procedures for learning rules for a given concept by generalizing from examples of that concept, lies in when the generalization is made. In fact, while CBR starts with a set of cases of training examples and forms generalizations of these examples by identifying commonalities between a retrieved case and the target problem, a rule-induction algorithm draws generalizations before the target problem is even known, i.e. it performs eager generalization. In mathematics, for example, the process of proving the truth of a proposition depending on a non negative integer by applying induction can be consolidated by generalizing a series of suitable examples, i.e. by a rule induction algorithm. On the contrary, when a concrete problem is given, the solver has simply to retrieve in memory an analogous problem solved in the past by induction and apply the same method for the solution of the given problem (CBR). The comparison between the CBR and the rule-based reasoning methodologies is presented in the following table:

**Table 1: Comparison between CBR and rule-based reasoning methodologies**

| Argument | Case-based | Rule-based |
|---|---|---|
| Knowledge source | Experience | Knowledge engineer. |
| The basic unit of | Case | Rule |
| knowledge. | | |
| Knowledge acquisition. | By assimilating new cases either first hand or through reports from others. | By adding new rules through knowledge engineer.(knowledge acquisition bottleneck). |
| Remembering | Can remember its own experience | Can't remember its experience |
| Learning | Can learn from his/her mistakes | Can't learn |
| Reasoning | Can reason by analogy | Can't reason by analogy. |

The CBR methodology directly addresses the following problems found also in rule-based technology.

1. *Knowledge acquisition:* The unit of knowledge is the case, not the rule. It is easier to articulate, examine, and evaluate cases than rules.
2. *Performance:* A CBR system can remember its own performance, and can modify its behavior to avoid repeating prior mistakes.
3. *Adaptive Solutions:* By reasoning from analogy with past cases, a CBR system should be able to construct solutions to novel problems.
4. *Maintaining:* Maintaining a CBR system is easier than a rule-based system since adding new knowledge can be as simple as adding a new case.

The idea of CBR is becoming popular in developing knowledge-based systems because it automates applications that are based on precedent or that contain incomplete causal models [87]. In a rule-based system an incomplete mode or an environment which does not take into account all variables could result in either an answer built on incomplete data or simply in no answer at all. The CBR methodology attempts to get around this shortcoming by inputting and analyzing problem data.

Research reveals that students learn best when they are presented with examples (cases) of problem-solving knowledge and then are required to apply this knowledge to real situations. The case-base of examples and exercises captures realistic problem-solving situations and presents them to the students as virtual simulations. Each example/exercise includes:

- A multi-media description of the problem, which may evolve over time.
- A description of the correct actions to take including order-independent, optional, and alternative steps.
- A multi-media explanation of why these steps are correct;
- The list of methods to determine whether students correctly executed the steps;
- The list of principles that must be learned to take the correct action.

All inductive reasoning, where data is too scarce for statistical relevance, is inherently based on anecdotal evidence. *Critics of CBR* argue that it is an approach that accepts anecdotal evidence as its main operating principle, but without statistically relevant data for backing an implicit generalization, there is no guarantee that the generalization is correct. Our personal opinion is that the above criticism has only a theoretical base, because in practice the CBR methods give satisfactory results in most cases.

Conclusively CBR has blown a fresh wind and a well justified degree of optimism into AI in general, and knowledge based decision support systems in particular. The growing amount of on going CBR research has the potential of leading into significant breakthroughs of AI methods and applications.

## References

[1] Aamodt A. (1989), Towards robust expert systems than learn from experience – an architectural framework, In Boose J., Gaines B., Ganascia J.- G. (Eds.): EKAW-89, *Third European Knowledge Acquisition for Knowledge-Based Systems Workshop*, 311-326, Paris.

[2] Aamodt A. (1993), Explanation-driven retrieval, reuse and learning of cases, In EWCBR-93: *First European Workshop on Case-Based Reasoning*, University of Kaiserslautern SEKI Report SR-93-12 (SFB 314), 279-284.

[3] Aamodt, A. & Plaza, E. (1994), Case-Based Reasoning:: Foundational Issues, Methodological Variations, and System Approaches, *A. I. Communications*, 7, no. 1, 39-52.

[4] Abdrabou, E. A. M. & Salem, A. B. (2009), Application of Case-Based Reasoning Frameworks in a Medical Classifier, *Proceedings of 4$^{th}$ Int., Conf. on Intelligent Computing and Information Systems*, Cairo, pp. 253-259.

[5] Abdrabou, E. A. M. & Salem, A. B. (2010), A Breast Cancer Classifier based on a Combination of Case-Based Reasoning and Ontology Approach, *Proc. of 2$^{nd}$ International Multi-conference on Computer Science and Information Technology ( IMCSIT 2010)*, Wisła, Poland.

[6] Aha, D,, Kibler, D. & Albert, M. K. (1991), Instance-Based Learning Algorithms, *Machine Learning*, Vol. 6 (1).

[7] *AI-CBR Website*: www.ai-cbr.org , Department of Computer Science, University of Auckland.

[8] Althoff K. D. (1989), Knowledge acquisition in the domain of CNC machine centers: the MOLTKE approach, In Boose J. Gaines B., Ganaskia J.-G. (Eds.): EKAW-89, *Third European Knowledge Acquisition for Knowledge-Based Systems Workshop*, 180-195, Paris.

[9] *Artificial Intelligence Applications Institute (AIAI) Website*: www.aiai.ed.ac.uk , School of Informatics, University of Edinburgh.

[10] Ashley K. (1991), *Modeling legal arguments: Reasoning with cases and hypotheticals*, MIT Press, Bradford Books, Cambridge

[11] Astin, A. W. (1993), *What matters in college? Four critical years revisited*, Jossey-Bass Inc., San Francisco.

[12] Bazzini, L. (1997), Revisiting analogy in learning mathematics, *Proceedings of the 1$^{st}$ Mediterranean Conference on Mathematics*, 174-181, Cyprus.

[13] Bareiss, R. (1989), *Exemplar-based knowledge acquisition: A unified approach to concept representation, classification, and learning*, Boston, Academic Press.

[14] Bower, G., Black, J. and Turner, T. (1979), Scripts in Memory for Text, *Cognitive Psychology*, 11, 177-220.

[15] Brown, B. , Lewis, L. (1991), A case-based reasoning solution to the problem of redundant resolutions of non-conformances in large scale manufacturing, In: Smith R., Scott C. (Eds), *Innovative Applications for Artificial Intelligence 3*, MIT Press.

[16] *CBR Group Web server*:  www.cs.umass.edu/~cbr/index.html

[17] Collins, A. &  Quilliam, M. (1969), Retrieval Time from Semantic Memory, *Journal of Verbal Learning and Verbal Behavior*, 8, 240-247.

[18] Cooper, G.  &  Sweller, J. (1987), The effects of schema acquisition and rule automation on  mathematics problem solving transfer, *J. of Educational Psychology*, 79, 347-362.

[19] Dunker, K. (1945), On problem solving, *Psychological Monographs*, 58 (5, whole No.  270).

[20] Feigenbaum, E. A. & McCorduck, P. (1983), *The fifth generation* (1st Ed.), Reading, MA: Addison-Wesley.

 [21] Fischbein, E., (1989), Tacit models and mathematical reasoning, *For the Learning of Mathematics*, 2, 9-14.

[22] Gelheiser, L. (1984), Generalization from categorical memory tasks to prose learning  in learning disable adolescents, *J. of Educational Psychology*, 76, 1128-1138 .

[23] Gentner, D. (1983), Structure mapping – a theoretical framework for analogy, Cognitive Science, 7, 155-170

[24] Genter, D. & Toupin, C. (1986), Systematicity and surface similarity in development of   analogy, *Cognitive Science*, 10, 277-300.

[25] Gick, M. L. & Holyoak K.  J. (1980),  Analogical problem solving, *Cognitive Psychology*, 12, 306-355.

[26] Gick, M. L. &  Holyoak, K. J. (1983), Schema induction and analogical transfer, *Cognitive Psychology*,  15, 1-38.

[27] Hall, R. P. (1989), Computational approaches to analogical reasoning: A comparative analysis, *Artificial Intelligence*, 39 (1), 39-120

[28] Halpern, D. F. (1997), *Critical thinking across the curriculum: A brief edition of thought and knowledge*, Lawrence Erlbaum associates, London,

[29] Hans-Dieter, Salem A. B., El Bagoury, B. M. (2007), Ideas of Case-Based Reasoning for Keyframe Technique, *Proceedings of the 16th International Workshop on the Concurrency Specification and Programming*,,Logow, Warsa, Poland, pp. 100-106.

[30] Harmon, P. (1992), *Case-based reasoning III*, *Intelligent Software strategies VIII(1)*.

[31] Hayes, J. R. & Simon, H. A. (1977), Psychological differences among problem isomorphs. In: N. J. Castellan, D. B. Pisoni and G. R. Potts (Eds.), *Cognitive Theory*, Vol. 2, 21-44, Hillsdale, NJ: Erlbaum.

[32] Heindl, B. et al, (1997), A Case-Based Consiliarius for Therapy Recommendation (ICONS) computer-based advise for calculated antibiotic therapy in intensive care medicine, *Computer Methods and Programs in Biomedicine*, 52, 117-127.

[33] Hennessy, D., and Hinkle, D. (1992), Applying Case-based reasoning to autoclave loading, *IEEE Expert*, 7(5), 21-26.

[34] Hinkle, D. & Toomey, C. (1995), Applying Case-Based Reasoning to Manufacturing, *AI Magazine*, 65-73.

[35] Holyoak, K. J. (1985), The pragmatics of analogical transfer. In: G. H. Bower (Ed.), *The psychology of learning and motivation,* Vol. 19, 59-87, New York: Academic Press.

[36] Holyoak, K. J. & Koh, K. (1987), Surface and structural similarity in analogical transfer, *Memory and Cognition*, 15, 332-340.

[37] Kay, D.S. & Black, J. B. (1985), The evolution of knowledge representations with increasing expertise in using systems, *Proceedings of the 7th Annual Conference of the Cognitive Science Society*, 140-149.

[38] Keane, M. (1988), Where's the Beef? The Absence of Pragmatic Factors in Pragmatic Theories of Analogy, *Proc. ECAI-88*, 327-333

[39] Kedar-Cabelli, S. (1988), Analogy – from a unified perspective. In Helman, D. H. (Ed.), *Analogical reasoning*, 65-103, Kluwer Academic.

[40] Kemeny, J. & Snell, J. l. (1976), *Finite Markov Chains*, Springer-Verlag, New York.

[41] Kibler, D. & Aha, D. (1987) Learning representative exemplars of concepts: An initial study, *Proceedings of the 4$^{th}$ International Workshop on Machine Learning*, UC- Irving, 24-29.

[42] Kintsch, W. (1972), Notes on the Structure of Semantic Memory. In: Organization of Memory, E. Tulvin and W. Donaldson (Eds.), 247-308, New York, Academic.

[43] Kitano, H. (1993), Challenges for massive parallelism, *Proceedings of the 13$^{th}$ Int. Conference on AI*, 813-834, Morgan Kaufman, Chambery, France.

[44] Klir, G. J. & Folger, T. A. (1988), *Fuzzy Sets, Uncertainty and Information*, Prentice Hall Int., London

[45] Kolodner, J. (1983), Reconstructive Memory: A Computer Model, *Cognitive Science*, 7, 281-328.

[46] Kolodner, J. (1988), Retrieving events from case memory: A parallel implementation, *Proceedings from the Case-based Reasoning Workshop*, 233-249, Morgan Kaufmann Publ., Clearwater Beach, Florida.

[47] Kolodner, J. L. (1991), Improving human decision making through Case-based decision aiding, *AI Magazine*, 52-69.

[48] Kolodner, J. (1993), *Case-Based Reasoning*, Morgan Kaufmann

[49] Koton, Ph. (1989), *Using experience in learning and problem solving*, MIT, Laboratory of Computer Science (Ph.D. Thesis.), MIT/LCS/TR-441

[50] Lebowitz, M. (1983), Memory-Based Parsing, *Artificial Intelligence*, 21, 363-404.

[51] Lei, Y., Peng, Y. & Ruan, X. (2001), Applying case'-based reasoning to cold forcing process planning, *Journal of Materials Processing Technology*, 112, 12-16.

[52] Lenz, M., Wess, S., Burkhard, H. & Bartsch, B. (1998), *Case based Reasoning technology: From foundations to applications*, Springer.

[53] Lopez, B. & Plaza E. (1993), Case-based planning for medical diagnosis. In: Kmorowski Z., Ras W. (Eds.), *Methodologies for Intelligence Systems*: 7$^{th}$

International Symposium (ISMIS 93), 96-105, Lecture Notes in Artificial Intelligence 689, Springer-Verlag.

[54] *Machine learning network on line information service*, available on the Web at www.kdubig.org/kdubig/control/index

[55] Manago, M., et al (1993), Induction and reasoning from cases. In: *ECML-European Conference on Machine Learning, Workshop on Intelligent Learning Architectures*, Vienna.

[56] Mark, B. (1989), Case-Based Reasoning for Autoclave Management, *Proceedings of the Case-Based Reasoning Workshop*.

[57] Mason, L. (1992), *Reti di somiglianze*, Franco Angeli .

[58] Michalski, R. & Tecuci, G. (1992), *Proc. Multi-strategy Learning Workshop*, George Mason University.

[59] Minsky, M. (1975), A Framework for Representing Knowledge. In: *The Psychology of Computer Vision*, Wilson, P. (Ed.), 211-277, New York, McGraw-Hill.

[60] National Council of Teachers of Mathematics (2010), *Principles and standards of school mathematics*, available at http://standards.nctm.org .

[61] Nguyen, T., Czerwinski, M. & Lee, D. (1993), COMPAQ Quick Source: Providing the Consumer with the Power of Artificial Intelligence, *Proceedings of the $5^{th}$ Annual Conference on Innovative Applications of Artificial Intelligence*, 142-151, AAAI Press, Washington DC.

[62] Novick, L. R. (1988), Analogical transfer, problem similarity and expertise, *Journal of Educational Psychology: Learning, Memory and Cognition*, 14, 510-520.

[63] *Official server of the International Conference on CBR*, available at www.iccbr.org .

[64] Pascarella, E. T. & Terenzini, P. (1991) *How college affects students*, Jossey-Bass, San Francisco.

[65] Plaza, E. & Lopez de Mantaras, R. (1990), A case-based apprentice that learns from fuzzy examples. In: Ras Z., Zemankova M., Emrich M. L. (Eds.), *Methodologies for Intelligent System*, 5, 420-427, North Holland

[66] Plaza, E. & Arcos, J. L. (1993), Reflection and Analogy in Memory-based Learning, *Proc. Multi-strategy Learning Workshop*, 42-92.

[67] Polya, G. (1945), *How to solve it,* Princeton Univ. Press, Princeton.

[68] Polya, G. (1954). *Mathematics and Plausible Reasoning*, Princeton Univ. Press, Princeton.

[69] Polya, G. (1954), *Induction and Analogy in Mathematics*, Princeton University Press, Princeton .

[70] Polya, G. (1963), On learning, teaching and learning teaching, *American Mathematical Monthly,* 70, 605-619.

[71] Polya, G. (1962/65), *Mathematical Discovery* (2 Volumes), J.Wilet & Sons, New York.

[72] Porter, B. & Bareiss, B. (1986), PROTOS: An experiment in knowledge acquisition for heuristic classification tasks, *Proceedings of the $1^{st}$ Intern. Meeting on Advances in Learning* (IMAL), 159-174, Les Arcs, France.

[73] Quilliam, M. (1968), Semantic Memory. In: *Semantic Information Processing*, Minsky, M. (Ed), 227-353, Cambridge, Mass., MIT Press.

[74] Recio-García, J., A. Bridge, D., Díaz-Agudo, B. 7González-Calero, P. (2008), A. CBR for CBR: A Case-Based Template Recommender System. In K.-D. Althoff and R. Bergmann (eds.), *Advances in Case-Based Reasoning*, 9th European Conference, LNCS, Springer.

[75] Richter, A. M. & Weiss, S. (1991), Similarity, uncertainty and case-based reasoning in PATDEX. In: Boyer R. S. (Ed.), *Automated reasoning essays in honour of Woody Bledsoe*, 249-265, Kluwer.

[76] Riesbeck, C. & Bain, W. (1987), *A Methodology for Implementing Case-Based Reasoning Systems*, Lockheed.

[77] Riesbeck, C. & Schank, R. (1989), *Inside case-based reasoning*, Lawrence Erlbaum.

[78] Rissland, E. (1983), Examples in legal reasoning: Legal hypotheticals. In *Proceedings of the $8^{th}$ International Joint Conference on Artificial Intelligence* (IJCAI), Karlsruhe

[79] Rissland, E. L. & Danials, J. J. (1995), A Hybrid CBR-IR Approach to Legal Information Retrieval, *Proceedings of the Fifth International Conference on Artificial Intelligence and Law (ICAIL-95)*, pp. 52-61, College Park, MD,.

[80] Ross, B. H. (1984), Remindings and their effects in learning a cognitive skill, *Cognitive Psychology*, 16, 371-416.

[81] Roth-Berghofer, T., R. & Bahls, D. (2008), *Explanation Capabilities of the Open Source Case-Based Reasoning*, Tool my CBR.

[82] Rumelhart, D., Lindsay, P. & Norman, D. (1972), A Process Model for Long-Term Memory. In: *Organization of Memory*, Tulving & Donadlson (Eds.), 197-246, New York, Academic.

[83] Salem, A-B. M. & Baeshen, N. (1999), Artificial Intelligence Methodologies for Developing Decision Aiding Systems, *Proceedings of Decision Sciences Institute, $5^{th}$ International Conference, Integrating Technology and Human Decisions: Global Bridges into the $21^{st}$ Century (D.I.S. 99 Athens),* Greece, pp.168-170,

[84] Salem, A-B. M., Roushdy M., & El-Bagoury, B.M. (2001), An Expert System for Diagnosis of Cancer Diseases, *Proceedings of the $7^{th}$ International Conference on Soft Computing (MENDEL)*, pp. 300-305.

[85] Salem, A-B. M. & HodHod, R. A. (2002), A Hybrid Expert System Supporting Diagnosis of Heart Diseases, , Kluwer Academic Publishers, *Proceedings of IFIP 17th World Computer Congress, TC12 Stream on Intelligent Information Processing* Montreal, Quebec, Canada, pp. 301-305.

[86] Salem, A.-B. M. , Bassant, M. & El Bagoury (2003), A Case-Based Adaptation Model for Thyroid Cancer Diagnosis Using Neural Networks, *Proceedings of the sixteenth international FLAIRS Conference*, AAAI Press, pp.155-159.

[87] Salem, A-B. M. (2007), Case Based Reasoning Technology for Medical Diagnosis**,** *Proceedings of World Academy of Science, Engineering And Technology (CESSE)*, Venice, Italy, Vol. 25, pp. 9-13.

[88] Salem, A.-B. M. (2012), Machine Learning in Biomedical informatics, Plenary lecture, *WSEAS $16^{th}$ International Conference on Computers*, July 18-17, Kos island, Greece.

Voskoglou & Salem

[89] Salomon, G. & Perkins, D. (1989), Rocky roads to transfer: Rethinking mechanism of a neglected phenomenon, *Educational Psychologist*, 24, 113-142.

[90] Schank, R. (1975), The Structure of Episodes in Memory. In: *Representation and Understanding*, Bobrow, D., G. & Collins, A. (Eds.), 237-272, New York, Academic.

[91] Schank, R. & Abelson, R. (1977), *Scripts, Plans, Goals and Understanding*, Hillsdale, N. J., Lawrence Erlbaum.

[92] Schank, R. (1979), *Reminding and Memory Organization: An Introduction to MOPS*, Technical Report 170, Dept. of Computer Science, Yale University.

[93] Schank, R. (1980), Language and Memory, *Cognitive Science*, 4(3), 243-284.

[94] Schank, R. (1981), Failure Driven Memory, *Cognition and Brain Theory*, 4(1), 41-60.

[95] Schank, R. (1982), *Dynamic memory: A theory of reminding and learning in computers and people,* Cambridge Univ. Press.

[96] Schank, R. & Leake D. (1989), Creativity and learning in case-based explainer, *Artificial Intelligence*, 40, no. 1-3, 353-385.

[97] Schoenfeld, A. H. & Herman, D. J. (1982), Problem perception and knowledge structure in expert and novice mathematical problem solvers, *J. of Experimental Psychology: Learning Memory and Cognition*, 8, 484-494 .

[98] Sharma, S. & Sleeman, D. (1988), REFINER: A case based differential diagnosis aide for knowledge acquisition and knowledge refinement. In EWSL 88: *Proceedings of the 3 $^d$ European Working Session on Learning*, 201-210, Pitman.

[99] Silvana, Q., Pedro, B., and Steen, A. (2001) *Proceedings of 8th Conference on Artificial Intelligence in Medicine in Europe*, AIME, Cascais, Portogal, Springer.
.
 [100] Skalak, C. B. & Rissland, E. (1992), Arguments and cases: An inevitable twining, *Artificial Intelligence and Law: An International Journal*, 1(1), 3-48.

[101] Slade, S. (1991), Case- Based Reasoning: A Research Paradigm, *Artificial Intelligence Magazine*, 12(1), 42-55.

[102] Smith E. & Medin, D. (1981), *Categories and concepts*, Harvard University Press.

[103] Spearman, C. (1923), *The nature of intelligence and the principles of cognition*, London: Macmillan.

[104] Stanfill, C. & Waltz, D. (1988), The memory-based reasoning paradigm, In: *Case-based reasoning, Proceedings from a workshop*, 414-424, Morgan Kaufmann Publ., Clearwater Beach, Florida.

[105] Steels, L. (1990), Components of expertise, *AI Magazine*, 11(2), 29-49.

[106] Steels, L. (1993), The componental framework and its role in reusability, In David J. M., Krivine, J. P. & Simmons R. (Eds.), *Second generation expert systems*, 273-298, Springer.

[107] Strube, G. & Janetzko, D. (1990), Epishodishes Wissen und Fallbasierte Schliessen: Aufgade fur die Wissendsdiagnostik und Wissenspsychologie, *Schweizerische Zeitschrift fur Psychologie*, 49, 211-221.

[108] Tulving, E. (1972), Episodic and semantic memory, In: Tulving, E. & Donaldson, W. (Eds.), *Organization of memory*, 381-403, Academic Press.

[109] Tulving, E. (1983), *Elements of Episodic Memory*, Oxford, Oxford University Press.

[110] Veloso, M. M. & Carbonell J. (1993), Derivational analogy in PRODIGY, *Machine Learning*, 10(3), 249-278.

[111] Venkatamaran, S. et al, (1993), A rule-rule case based system for image analysis, *1st European Workshop on Case-based Reasoning, Posters and Presentations*, Vol. II, 410-415, University of Kaiserslautern.

[112] Voskoglou, M. Gr. (1996), Use of absorbing Markov chains as a measurement model for the process of analogical transfer, *Int. J. of Mathematical Education in Science and Technology*, 27, 197-205.

[113] Voskoglou, M. Gr. (2003), Analogical problem solving and transfer, *Proceedings 3rd Mediterranean Conference in Mathematics Education*, 295-303, Athens, Greece.

[114] Voskoglou, M. Gr. (2007), The problem as a learning device of mathematics. In P. Avgerinos & A. Gagatsis (Eds.), *Current Trends in Mathematics Education*, 223-232, New Technologies Publications, Athens,.

[115] Voskoglou, M. Gr. (2008), Problem- Solving in Mathematics Education: Recent Trends and Development, *Quaderni di Ricerca in Didattica (Scienze Mathematiche)*, University of Palermo, 18, 22-28, 2008.

[116] Voskoglou, M. Gr. (2008), Case-Based Reasoning: A Recent Theory for Problem-Solving in Computers and People, *Communications in Computer and Information Science* (Springer), 19, 314-319.

[117] Voskoglou, M. Gr. (2009), Fuzzy sets in Case-Based Reasoning, *Fuzzy Systems and Knowledge Discovery*, 6 , 252-256, IEEE Computer Society.

[118] Voskoglou, M. Gr. (2010), A Stochastic Model for Case-Based Reasoning, *International Journal of Modelling and Application* (Univ. Blumenau, Brazil), 3, 33-39.

[119] Voskoglou, M. Gr. & Subbotin, I. (2012), Fuzzy Models for Analogical Reasoning, *International Journal of Applications of Fuzzy Sets and Artificial Intelligence*, 2, 19-38.

[120] Voss, A. (1996), Towards a Methodology for Case Adaptation, Proceedings of the 12th European Conference on Artificial Intelligence, Budapest, Hungary, pp. 147–157.

[121] Voss, J. F. (1987), Learning and transfer in subject matter learning: A problem-solving model, *International Journal of Educational Research*, 11, 607-622.

[122] Watson, I., (1997), *Applying Case-Based Reasoning: techniques for Enterprise Systems*, Elsevier: Burlington.

[123]Wittgenstein, L. (1955), Philosophical investigations, 31-34, Blackwell.

[124] Woods, W. (1975), What is a Link: Foundations for Semantic Networks. In: *Representation and Understanding*, Bobrow, D. & Collins, A. (Eds.), 35-82, New York, Academic.
.

**Authors' short Bio-Notes**

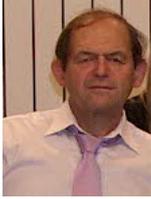

**Dr. Michael Gr. Voskoglou** is an Emeritus Professor of Mathematical Sciences at the Graduate Technological Educational Institute of Western Greece. He is the author of 9 books (8 in Greek and 1 in English language) and of more than 300 papers published in reputed journals and in proceedings of conferences of 25 countries in 5 continents, with many citations from other researchers. He is also the Editor- in- Chief of the "International Journal of Applications of Fuzzy Sets and Artificial Intelligence" (e-journal), a reviewer of the American Mathematical Society and member of the Editorial Board or referee in several mathematical journals. His research interests include Algebra, Markov Chains, Fuzzy Sets and Mathematics Education.

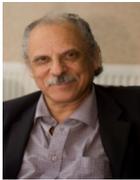

**Dr. Abdel-Badeeh M Salem** is a Professor of Computer Science since 1989 at Ain Shams University, Egypt. His research includes intelligent computing, knowledge-based systems, biomedical informatics, and intelligent e-learning. He has published around 250 papers in refereed journals and conferences. He has been involved in more than 400 conferences and workshops as a Keynote Speaker, Scientific Program Committee, Organizer and Session Chair. He is a member of many national and international informatics associations.